\documentclass{article}

\usepackage[utf8]{inputenc} 
\usepackage[T1]{fontenc}    
\usepackage{lmodern}
\usepackage{url}            
\usepackage{booktabs}       
\usepackage{amsfonts}       
\usepackage{nicefrac}       
\usepackage{microtype}      
\usepackage{setspace}
\usepackage{caption}
\usepackage{enumitem}
\usepackage{xfrac}
\usepackage{amsmath}
\usepackage[square,semicolon]{natbib}
\usepackage[backref=page]{hyperref}
\usepackage{xcolor}
\usepackage{rotating}
\usepackage[paper=8.5in:11in]{typearea}
\usepackage[margin=1.0in]{geometry}
\usepackage{amssymb}
\usepackage{comment}

\usepackage{multicol}
\usepackage{microtype}
\usepackage{graphicx}
\usepackage[procnames]{listings}
\definecolor{keywords}{RGB}{255,0,90}
\definecolor{comments}{RGB}{0,0,113}
\definecolor{red}{RGB}{160,0,0}
\definecolor{green}{RGB}{0,100,0}
 
\definecolor{darkblue}{rgb}{0, 0.2, 0.7}
\hypersetup{
    colorlinks = true,
    linkcolor = darkblue,
    anchorcolor = darkblue,
    citecolor = darkblue,
    filecolor = darkblue,
    urlcolor = darkblue
}

\title{\vspace{-2em}%
  \hrule height 4pt%
  \vskip 0.25in%
  \vskip -\parskip%
  \textbf{
  GLU Variants Improve Transformer 
  }%
  \vskip 0.2in%
  \vskip -\parskip%
  \hrule height 1pt%
  \vskip 0.09in}

\author{Noam Shazeer \\ Google \\ noam@google.com}

\begin{document}

\maketitle

\begin{abstract}
Gated Linear Units \citep{dauphin2016} consist of the component-wise product of two linear projections, one of which is first passed through a sigmoid function.  Variations on GLU are possible, using different nonlinear (or even linear) functions in place of sigmoid.  We test these variants in the feed-forward sublayers of the Transformer \citep{Vas17} sequence-to-sequence model, and find that some of them yield quality improvements over the typically-used ReLU or GELU activations.
\end{abstract}


\section{Introduction}
The Transformer \citep{Vas17} sequence-to-sequence model alternates between multi-head attention, and what it calls "position-wise feed-forward networks" (FFN).  The FFN takes a vector $x$ (the hidden representation at a particular position in the sequence) and passes it through two learned linear transformations, (represented by the matrices $W_1$ and $W_2$ and bias vectors $b_1$ and $b_2$).  A rectified-linear (ReLU) \citep{glorot2011deep} activation function applied between the two linear transformations.

\begin{equation}
\textrm{FFN}(x, W_1, W_2, b_1, b_2) = \textrm{max}(0, xW_1 + b_1)W_2 + b_2
\end{equation}

Following the T5 codebase \citep{raffel2019exploring} \footnote{Also in the interest of ML fairness.}, we use a version with no bias:

\begin{equation}
\textrm{FFN}_\textrm{ReLU}(x, W_1, W_2) = \textrm{max}(xW_1, 0)W_2
\end{equation}

Subsequent work has proposed replacing the ReLU with other nonlinear activation functions such as Gaussian Error Linear Units, $\textrm{GELU}(x)=x\Phi(x)$ \citep{hendrycks2016}, and $\textrm{Swish}_\beta(x)=x\sigma(\beta x)$ \citep{ramachandran2017searching}.

\begin{equation}
\begin{split}
\textrm{FFN}_\textrm{GELU}(x, W_1, W_2) & = \textrm{GELU}(xW_1)W_2 \\
\textrm{FFN}_\textrm{Swish}(x, W_1, W_2) & = \textrm{Swish}_1(xW_1)W_2
\end{split}
\end{equation}

\section{Gated Linear Units (GLU) and Variants}

\citep{dauphin2016} introduced Gated Linear Units (GLU), a neural network layer defined as the component-wise product of two linear transformations of the input, one of which is sigmoid-activated.  They also suggest omitting the activation, which they call a "bilinear" layer and attribute to \citep{mnih2007three}.

\begin{equation}
\begin{split}
\textrm{GLU}(x, W, V, b, c) & = \sigma(xW + b) \otimes (xV + c) \\
\textrm{Bilinear}(x, W, V, b, c) & = (xW + b) \otimes (xV + c)
\end{split}
\end{equation}

We can also define GLU variants using other activation functions:

\begin{equation}
\begin{split}
\textrm{ReGLU}(x, W, V, b, c) & = \textrm{max}(0, xW + b) \otimes (xV + c) \\
\textrm{GEGLU}(x, W, V, b, c) & = \textrm{GELU}(xW + b) \otimes (xV + c) \\
\textrm{SwiGLU}(x, W, V, b, c, \beta) & = \textrm{Swish}_\beta(xW + b) \otimes (xV + c) \\
\end{split}
\end{equation}

In this paper, we propose additional variations on the Transformer FFN layer which use GLU or one of its variants in place of the first linear transformation and the activation function.  Again, we omit the bias terms.

\begin{equation}
\begin{split}
\textrm{FFN}_\textrm{GLU}(x, W, V, W_2) & = (\sigma(xW) \otimes xV)W_2 \\
\textrm{FFN}_\textrm{Bilinear}(x, W, V, W_2) & = (xW \otimes xV)W_2 \\
\textrm{FFN}_\textrm{ReGLU}(x, W, V, W_2) & = (\textrm{max}(0, xW) \otimes xV)W_2 \\
\textrm{FFN}_\textrm{GEGLU}(x, W, V, W_2) & = (\textrm{GELU}(xW) \otimes xV)W_2 \\
\textrm{FFN}_\textrm{SwiGLU}(x, W, V, W_2) & = (\textrm{Swish}_1(xW) \otimes xV)W_2 \\
\end{split}
\end{equation}

All of these layers have three weight matrices, as opposed to two for the original FFN.  To keep the number of parameters and the amount of computation constant, we reduce the number of hidden units $d_{ff}$ (the second dimension of $W$ and $V$ and the first dimension of $W_2$) by a factor of $\frac{2}{3}$ when comparing these layers to the original two-matrix version.

\section{Experiments on Text-to-Text Transfer Transformer (T5)}

We test the FFN variants we have described on the transfer-learning setup from \citep{raffel2019exploring}.  An encoder-decoder transformer model \citep{Vas17} is trained on a denoising objective of predicting missing text segments, and subsequently fine-tuned on various language understanding tasks.  

\subsection{Model Architecture}

We use the same code base, model architecture, and training task as the base model from \citep{raffel2019exploring}.  The encoder and decoder each consist of 12 layers, with $d_{model}=768$.  For the attention layers,  $h=12$ and $d_k = d_v = 64$.   The FFN layers have hidden size $d_{ff}=3072$.  As we describe above, for the GLU-variant-based FFN layers, which have thee weight matrices instead of two, we reduce the hidden layer to $d_{ff}=2048$, so as to maintain the same parameter and operation counts as the base model.

\begin{table}[h]
\caption{Heldout-set log-perplexity for Transformer models on the segment-filling task from \citep{raffel2019exploring}.  All models are matched for parameters and computation.}
\label{tab:ppl}
\begin{center}
\vspace{-2mm}
\scalebox{1.0}{
\begin{tabular}{l|ll}
Training Steps & 65,536 & 524,288    \\
\hline
$\textrm{FFN}_\textrm{ReLU} (baseline)$  & 1.997 (0.005) & 1.677  \\
$\textrm{FFN}_\textrm{GELU}$  & 1.983 (0.005) & 1.679  \\
$\textrm{FFN}_\textrm{Swish}$  & 1.994 (0.003) & 1.683  \\
\hline
$\textrm{FFN}_\textrm{GLU}$  & 1.982 (0.006) & 1.663  \\
$\textrm{FFN}_\textrm{Bilinear}$  & 1.960 (0.005) & 1.648  \\
$\textrm{FFN}_\textrm{GEGLU}$  & \textbf{1.942} (0.004) & \textbf{1.633}  \\
$\textrm{FFN}_\textrm{SwiGLU}$  & \textbf{1.944} (0.010) & \textbf{1.636}  \\
$\textrm{FFN}_\textrm{ReGLU}$  & 1.953 (0.003) & 1.645  \\
\end{tabular}
}
\end{center}
\end{table}

\subsection{Pre-Training and Perplexity Results}
Identically to \citep{raffel2019exploring}, we pre-train for 524,288 steps on the span-filling objective on the C4 dataset.  Each training batch consists of 128 examples, each of which has an input of 512 tokens and an output of 114 tokens, the output containing multiple spans of tokens which were deleted from the input\footnote{Each training step took approximately 0.15 seconds on a 32-core TPUv2 cluster.}.  Similarly to \citep{raffel2019exploring}, we use the Adafactor optimizer \citep{shazeer2018adafactor} and an inverse-square-root learning-rate schedule.  We also decay the learning rate linearly for the final 10 percent of the training steps.   Our main departure from \citep{raffel2019exploring} is that we use no dropout during pre-training.  We find this to produce superior results. We compute the log-perplexity on the training objective on a heldout shard of C4, which we believe to be a good indicator of model quality.  For each model architecture, we also trained four models for a shorter period (65,536 steps) to measure inter-run variability.  The results are listed in table \ref{tab:ppl}.  The GEGLU and SwiGLU variants produce the best perplexities.

\subsection{Fine-Tuning}
We then fine-tune each fully-trained model once on an examples-proportional mixture of the Stanford Question-Answering Dataset (SQuAD) \citep{rajpurkar2016squad} and all the language understanding tasks in the GLUE \citep{wang2018glue} and SuperGlue \citep{wang2019superglue} benchmarks.\footnote{This departs from \citep{raffel2019exploring}, who fine-tuned separately on the different tasks.  We chose one fine-tuning run for simplicity.}  Fine-tuning consists of 131072 steps with a learning rate of $10^{-3}$.  As in training, the input sequences for each step have a combined length of approximately 65,536 tokens.  Following \citep{raffel2019exploring}, we use a dropout rate of $0.1$ on the layer outputs, feed-forward hidden-layers and attention weights.  The embedding matrices are fixed during fine-tuning.

Tables \ref{tab:glue}, \ref{tab:superglue} and \ref{tab:squad} show results on the development sets.  For each task, we report the best score of any of the checkpoints recorded during fine-tuning.  While the results are noisy, the new GLU-variants perform best on most of the tasks.  For comparison, at the bottom of each of the tables we list the reuslts from \citep{raffel2019exploring}.  The model is identical to our $\textrm{FFN}_\textrm{ReLU}$ model.  Their results are notably worse, which we believe was caused by their use of dropout during pre-training.  Also listed are the inter-run standard deviations measured by \citep{raffel2019exploring}.

\begin{table}[h]
\caption{GLUE Language-Understanding Benchmark \citep{wang2018glue} (dev).}
\label{tab:glue}
\begin{center}
\vspace{-2mm}
\scalebox{0.73}{
\begin{tabular}{l|c|cccccccccccc}
 & Score & CoLA & SST-2 & MRPC & MRPC & STSB & STSB & QQP & QQP & MNLIm & MNLImm & QNLI & RTE \\
 & Average & MCC & Acc &  F1 & Acc & PCC & SCC & F1 & Acc & Acc & Acc & Acc & Acc \\
 \hline
$\textrm{FFN}_\textrm{ReLU}$ &  $83.80$ &  $51.32$ &  $94.04$ &  \textbf{93.08} &  \textbf{90.20} &  $89.64$ &  $89.42$ &  $89.01$ &  $91.75$ &  $85.83$ &  $86.42$ &  $92.81$ &  $80.14$ \\
$\textrm{FFN}_\textrm{GELU}$ &  $83.86$ &  $53.48$ &  $94.04$ &  $92.81$ &  \textbf{90.20} &  $89.69$ &  $89.49$ &  $88.63$ &  $91.62$ &  $85.89$ &  $86.13$ &  $92.39$ &  $80.51$ \\
$\textrm{FFN}_\textrm{Swish}$ &  $83.60$ &  $49.79$ &  $93.69$ &  $92.31$ &  $89.46$ &  $89.20$ &  $88.98$ &  $88.84$ &  $91.67$ &  $85.22$ &  $85.02$ &  $92.33$ &  $81.23$ \\
\hline
$\textrm{FFN}_\textrm{GLU}$ &  $84.20$ &  $49.16$ &  $94.27$ &  $92.39$ &  $89.46$ &  $89.46$ &  $89.35$ &  $88.79$ &  $91.62$ &  $86.36$ &  $86.18$ &  $92.92$ &  \textbf{84.12} \\
$\textrm{FFN}_\textrm{GEGLU}$ &  $84.12$ &  $53.65$ &  $93.92$ &  $92.68$ &  $89.71$ &  $90.26$ &  $90.13$ &  $89.11$ &  $91.85$ &  $86.15$ &  $86.17$ &  $92.81$ &  $79.42$ \\
$\textrm{FFN}_\textrm{Bilinear}$ &  $83.79$ &  $51.02$ &  \textbf{94.38} &  $92.28$ &  $89.46$ &  $90.06$ &  $89.84$ &  $88.95$ &  $91.69$ &  \textbf{86.90} &  \textbf{87.08} &  $92.92$ &  $81.95$ \\
$\textrm{FFN}_\textrm{SwiGLU}$ &  $84.36$ &  $51.59$ &  $93.92$ &  $92.23$ &  $88.97$ &  \textbf{90.32} &  \textbf{90.13} &  \textbf{89.14} &  \textbf{91.87} &  $86.45$ &  $86.47$ &  \textbf{92.93} &  $83.39$ \\
$\textrm{FFN}_\textrm{ReGLU}$ &  \textbf{84.67} &  \textbf{56.16} &  \textbf{94.38} &  $92.06$ &  $89.22$ &  $89.97$ &  $89.85$ &  $88.86$ &  $91.72$ &  $86.20$ &  $86.40$ &  $92.68$ &  $81.59$ \\
\hline \hline
\citep{raffel2019exploring} & $83.28$ & $53.84$ & $92.68$ & $92.07$ & $88.92$ & $88.02$ & $87.94$ & $88.67$ & $91.56$ & $84.24$ & $84.57$ & $90.48$ & $76.28$ \\
ibid. stddev. & $0.235$ & $1.111$ & $0.569$ & $0.729$ & $1.019$ & $0.374$ & $0.418$ & $0.108$ & $0.070$ & $0.291$ & $0.231$ & $0.361$ & $1.393$ \\

\end{tabular}
}
\end{center}
\end{table}

\begin{table}[h]
\caption{SuperGLUE Language-Understanding Benchmark \citep{wang2019superglue} (dev).}
\label{tab:superglue}
\begin{center}
\vspace{-2mm}
\scalebox{0.75}{
\begin{tabular}{l|c|cccccccccccc}
 & Score & BoolQ & CB  & CB & CoPA & MultiRC & MultiRC & ReCoRD & ReCoRD  & RTE & WiC & WSC \\
 & Average & Acc & F1 & Acc & Acc & F1 & EM & F1 & EM & Acc & Acc & Acc \\
\hline
$\textrm{FFN}_\textrm{ReLU}$ &  $72.76$ &  $80.15$ &  $83.37$ &  $89.29$ &  $70.00$ &  $76.93$ &  $39.14$ &  $73.73$ &  $72.91$ &  $83.39$ &  $67.71$ &  $77.88$ \\
$\textrm{FFN}_\textrm{GELU}$ &  $72.98$ &  $80.64$ &  $86.24$ &  \textbf{91.07} &  $74.00$ &  $75.93$ &  $38.61$ &  $72.96$ &  $72.03$ &  $81.59$ &  $68.34$ &  $75.96$ \\
$\textrm{FFN}_\textrm{Swish}$ &  $72.40$ &  $80.43$ &  $77.75$ &  $83.93$ &  $67.00$ &  $76.34$ &  $39.14$ &  $73.34$ &  $72.36$ &  $81.95$ &  $68.18$ &  $81.73$ \\
\hline
$\textrm{FFN}_\textrm{GLU}$ &  $73.95$ &  $80.95$ &  $77.26$ &  $83.93$ &  $73.00$ &  $76.07$ &  $39.03$ &  $74.22$ &  $73.50$ &  $84.12$ &  $67.71$ &  \textbf{87.50} \\
$\textrm{FFN}_\textrm{GEGLU}$ &  $73.96$ &  $81.19$ &  $82.09$ &  $87.50$ &  $72.00$ &  \textbf{77.43} &  \textbf{41.03} &  $75.28$ &  \textbf{74.60} &  $83.39$ &  $67.08$ &  $83.65$ \\
$\textrm{FFN}_\textrm{Bilinear}$ &  $73.81$ &  \textbf{81.53} &  $82.49$ &  $89.29$ &  \textbf{76.00} &  $76.04$ &  $40.92$ &  $74.97$ &  $74.10$ &  $82.67$ &  \textbf{69.28} &  $78.85$ \\
$\textrm{FFN}_\textrm{SwiGLU}$ &  \textbf{74.56} &  $81.19$ &  $82.39$ &  $89.29$ &  $73.00$ &  $75.56$ &  $38.72$ &  \textbf{75.35} &  $74.55$ &  \textbf{85.20} &  $67.24$ &  $86.54$ \\
$\textrm{FFN}_\textrm{ReGLU}$ &  $73.66$ &  $80.89$ &  \textbf{86.37} &  \textbf{91.07} &  $67.00$ &  $75.32$ &  $40.50$ &  $75.07$ &  $74.18$ &  $84.48$ &  $67.40$ &  $79.81$ \\
\hline
\hline

\citep{raffel2019exploring} & $71.36$ & $76.62$ & $91.22$ & $91.96$ & $66.20$ & $66.13$ & $25.78$ & $69.05$ & $68.16$ & $75.34$ & $68.04$ & $78.56$ \\
ibid. stddev. & $0.416$ & $0.365$ & $3.237$ & $2.560$ & $2.741$ & $0.716$ & $1.011$ & $0.370$ & $0.379$ & $1.228$ & $0.850$ & $2.029$ \\

\end{tabular}
}
\end{center}
\end{table}

\begin{table}[h]
\caption{SQuAD \citep{rajpurkar2016squad} v1.1 (dev).}
\label{tab:squad}
\begin{center}
\vspace{-2mm}
\scalebox{0.75}{
\begin{tabular}{l|ll}
& EM & F1 \\
\hline
$\textrm{FFN}_\textrm{ReLU}$ &  $83.18$ &  $90.87$ \\
$\textrm{FFN}_\textrm{GELU}$ &  $83.09$ &  $90.79$ \\
$\textrm{FFN}_\textrm{Swish}$ &  $83.25$ &  $90.76$ \\
\hline
$\textrm{FFN}_\textrm{GLU}$ &  $82.88$ &  $90.69$ \\
$\textrm{FFN}_\textrm{GEGLU}$ &  $83.55$ &  $91.12$ \\
$\textrm{FFN}_\textrm{Bilinear}$ &  \textbf{83.82} &  $91.06$ \\
$\textrm{FFN}_\textrm{SwiGLU}$ &  $83.42$ &  $91.03$ \\
$\textrm{FFN}_\textrm{ReGLU}$ &  $83.53$ & \textbf{91.18} \\

\hline
\hline
\citep{raffel2019exploring} & $80.88$ & $88.81$ \\
ibid. Standard Deviation & $0.343$ & $0.226$ \\
\end{tabular}
}
\end{center}
\end{table}

\section{Conclusions}
We have extended the GLU family of layers and proposed their use in Transformer.  In a transfer-learning setup, the new variants seem to produce better perplexities for the de-noising objective used in pre-training, as well as better results on many downstream language-understanding tasks.  These architectures are simple to implement, and have no apparent computational drawbacks.  We offer no explanation as to why these architectures seem to work; we attribute their success, as all else, to divine benevolence.

\bibliography{main}

\begin{thebibliography}{11}
\providecommand{\natexlab}[1]{#1}
\providecommand{\url}[1]{\texttt{#1}}
\expandafter\ifx\csname urlstyle\endcsname\relax
  \providecommand{\doi}[1]{doi: #1}\else
  \providecommand{\doi}{doi: \begingroup \urlstyle{rm}\Url}\fi

\bibitem[Dauphin et~al.(2016)Dauphin, Fan, Auli, and Grangier]{dauphin2016}
Yann~N. Dauphin, Angela Fan, Michael Auli, and David Grangier.
\newblock Language modeling with gated convolutional networks.
\newblock \emph{CoRR}, abs/1612.08083, 2016.
\newblock URL \url{http://arxiv.org/abs/1612.08083}.

\bibitem[Glorot et~al.(2011)Glorot, Bordes, and Bengio]{glorot2011deep}
Xavier Glorot, Antoine Bordes, and Yoshua Bengio.
\newblock Deep sparse rectifier neural networks.
\newblock In \emph{Proceedings of the fourteenth international conference on
  artificial intelligence and statistics}, pages 315--323, 2011.

\bibitem[Hendrycks and Gimpel(2016)]{hendrycks2016}
Dan Hendrycks and Kevin Gimpel.
\newblock Bridging nonlinearities and stochastic regularizers with gaussian
  error linear units.
\newblock \emph{CoRR}, abs/1606.08415, 2016.
\newblock URL \url{http://arxiv.org/abs/1606.08415}.

\bibitem[Mnih and Hinton(2007)]{mnih2007three}
Andriy Mnih and Geoffrey Hinton.
\newblock Three new graphical models for statistical language modelling.
\newblock In \emph{Proceedings of the 24th international conference on Machine
  learning}, pages 641--648, 2007.

\bibitem[Raffel et~al.(2019)Raffel, Shazeer, Roberts, Lee, Narang, Matena,
  Zhou, Li, and Liu]{raffel2019exploring}
Colin Raffel, Noam Shazeer, Adam Roberts, Katherine Lee, Sharan Narang, Michael
  Matena, Yanqi Zhou, Wei Li, and Peter Liu.
\newblock Exploring the limits of transfer learning with a unified text-to-text
  transformer.
\newblock \emph{arXiv e-prints}, 2019.

\bibitem[Rajpurkar et~al.(2016)Rajpurkar, Zhang, Lopyrev, and
  Liang]{rajpurkar2016squad}
Pranav Rajpurkar, Jian Zhang, Konstantin Lopyrev, and Percy Liang.
\newblock Squad: 100,000+ questions for machine comprehension of text.
\newblock \emph{arXiv preprint arXiv:1606.05250}, 2016.

\bibitem[Ramachandran et~al.(2017)Ramachandran, Zoph, and
  Le]{ramachandran2017searching}
Prajit Ramachandran, Barret Zoph, and Quoc~V Le.
\newblock Searching for activation functions.
\newblock \emph{arXiv preprint arXiv:1710.05941}, 2017.

\bibitem[Shazeer and Stern(2018)]{shazeer2018adafactor}
Noam Shazeer and Mitchell Stern.
\newblock Adafactor: Adaptive learning rates with sublinear memory cost.
\newblock \emph{arXiv preprint arXiv:1804.04235}, 2018.

\bibitem[Vaswani et~al.(2017)Vaswani, Shazeer, Parmar, Uszkoreit, Jones, Gomez,
  Kaiser, and Polosukhin]{Vas17}
Ashish Vaswani, Noam Shazeer, Niki Parmar, Jakob Uszkoreit, Llion Jones,
  Aidan~N. Gomez, Lukasz Kaiser, and Illia Polosukhin.
\newblock Attention is all you need.
\newblock In \emph{NIPS}, 2017.

\bibitem[Wang et~al.(2018)Wang, Singh, Michael, Hill, Levy, and
  Bowman]{wang2018glue}
Alex Wang, Amapreet Singh, Julian Michael, Felix Hill, Omer Levy, and Samuel~R.
  Bowman.
\newblock {GLUE}: A multi-task benchmark and analysis platform for natural
  language understanding.
\newblock \emph{arXiv preprint arXiv:1804.07461}, 2018.

\bibitem[Wang et~al.(2019)Wang, Pruksachatkun, Nangia, Singh, Michael, Hill,
  Levy, and Bowman]{wang2019superglue}
Alex Wang, Yada Pruksachatkun, Nikita Nangia, Amanpreet Singh, Julian Michael,
  Felix Hill, Omer Levy, and Samuel~R. Bowman.
\newblock Superglue: A stickier benchmark for general-purpose language
  understanding systems.
\newblock \emph{arXiv preprint arXiv:1905.00537}, 2019.

\end{thebibliography}
\bibliographystyle{plainnat}


\end{document}